\documentclass[twocol]{ametsocv6}

\usepackage{amsmath,amsfonts,amssymb,bm}
\usepackage{mathptmx}
\usepackage{newtxtext}
\usepackage{newtxmath}

\usepackage{algorithm}
\usepackage[noend]{algpseudocode}
\usepackage{amsmath,amssymb,bbm}
\usepackage{blindtext}
\usepackage{booktabs}
\usepackage{blindtext}
\usepackage{graphicx}
\usepackage{graphics}
\usepackage{epstopdf}
\usepackage{mathtools}
\usepackage{rotating}
\usepackage{tikz}
\usepackage{adjustbox}
\usepackage[unicode=true,
 bookmarks=true,bookmarksnumbered=false,bookmarksopen=false,
 breaklinks=false,pdfborder={0 0 0},pdfborderstyle={},backref=false,colorlinks=true]
 {hyperref}
\usepackage{color}

\usepackage{scalerel,stackengine}
\stackMath
\newcommand\reallywidehat[1]{%
\savestack{\tmpbox}{\stretchto{%
  \scaleto{%
    \scalerel*[\widthof{\ensuremath{#1}}]{\kern-.6pt\bigwedge\kern-.6pt}%
    {\rule[-\textheight/2]{1ex}{\textheight}}
  }{\textheight}%
}{0.5ex}}%
\stackon[1pt]{#1}{\tmpbox}%
}
\bibpunct{(}{)}{;}{a}{}{,}

\title{A Deep Learning Approach to Probabilistic Forecasting of Weather}

\authors{Nick Rittler\correspondingauthor{Nick Rittler, nrittler@ucsd.edu}, Carlo Graziani \author{Carlo Graziani, cgraziani@anl.gov}, Jiali Wang, Rao Kotamarthi}

\affiliation{Argonne National Laboratory} 

\abstract{
We discuss an approach to probabilistic forecasting based on two chained
machine-learning steps: a \textit{dimensional reduction} step that
learns a reduction map of predictor information to a low-dimensional space in a
manner designed to preserve information about forecast quantities; and a
\textit{density estimation} step that uses the probabilistic machine learning
technique of normalizing flows to compute the joint probability density of
reduced predictors and forecast quantities. This joint density is then
renormalized to produce the conditional forecast distribution. In this method,
probabilistic calibration testing plays the role of a regularization procedure,
preventing overfitting in the second step, while effective dimensional reduction
from the first step is the source of forecast sharpness.  We verify the method
using a 22-year 1-hour cadence time series of Weather Research and Forecasting
(WRF) simulation data of surface wind on a grid.
}

\begin{document}

\maketitle

\section{Introduction} 
Weather forecast capabilities are crucial to energy market operations, regulation, and policymaking, as well as infrastructure maintenance and planning. Probabilistic forecasting of weather in particular \citep{palmer2002econ,zhu2002econ,gneiting2008surfacewind} is widely accepted and preferred in practice, largely due to the fact that a probabilisitic forecast, i.e. a \textit{distribution} over future outcomes, captures the uncertainty of the situation in a way that simple ``point'' predictions fail to do. However, limitations in physics and the need for higher spatial resolutions have limited the use of probabilistic forecasting for variables such as surface temperature, wind speeds and other weather variables of interest \citep{Hirschberg2011}. Thus, probabilistic forecasting still presents an important, and challenging problem.

The growing interest in machine learning in the past decade has resulted in a corresponding growth in machine learning-based weather forecasting literature. We observe, however, that modeling
 weather events in this literature has often focused on the reduction of forecasting tasks to classification 
 or point prediction settings, be it through fully data-driven techniques, or via the application of machine 
 learning techniques as a post-processing step following physics-based modeling.  Papers in the former category 
 include \cite{jergensen2020}, \cite{sprenger2017},  \cite{fabbian2007}, and
 \cite{roebber2007}; papers in the later category include \cite{yuan2007},  \cite{burke2020}, and 
 \cite{schaumann2021}.
  
 Interestingly, this focus is somewhat in contrast with the probabilistic forecasting community at large, wherein 
 there has been significant work on the application of machine learning to produce probabilisitic forecasts \citep{shi2018machineLF,lim2020deepforecast}. One of most common general approaches in this literature is the statistical learning of complex functions taking in current information and outputting parameters governing a distribution over future outcomes; the modeling of these functions has been done 
 by methods that include recurrent neural networks  \citep{salinas2020}, 
 convolutional neural networks  \citep{chen2020}, and standard fully connected nets \citep{vossen2018}. 

Research on machine learning of probabilistic forecasts for weather is now beginning to appear. \cite{petersik2020}, use deep neural networks to learn 
a nonlinear transformation of the current weather conditions into the parameters of a Gaussian mixture
to model the El Niño–Southern Oscillation. \cite{grover2015} produce point predictions 
of weather conditions at specific locales using standard procedures, use kernel methods
to enforce a smooth interpolation between locales, and take a Bayesian approach to produce estimates 
for the joint posterior distribution of weather conditions across locales.

Our approach to the probabilistic weather forecasting problem is to first learn an explicit dimension reduction, and 
thus a low dimensional representation of a large spatio-temporal grid of weather measurements, something that 
is conceptually not all that different from the learning of a transformation of input into distributional parameters. 
However, instead of enforcing a parametric assumption in the modeling of a future weather event, we non-parametrically
learn the distributional relationship of our low dimensional representation with the future outcomes. Until the last few 
years, procedures in the ML weather forecasting literature have largely relied on rather subjective selection of forecasting
 predictors from available data, often ignoring the potential that spatio-temporal grids of observations and other high dimensional data sets have for improving
forecast quality; though there have been recent efforts to create models with higher dimensional inputs  \citep{qui2017,mehrkanoon2019}, and though there are notable early exceptions \citep{nandar2009}, we 
believe our emphasis on the learning of a low dimensional representation of high dimensional weather 
data to be both an effective method and a useful illustration of the use of high dimensional data. Given the 
growth in simulation and observational data sample sizes, the joint modeling of low dimensional representations of the past and possible
 future observations through non-parametric methods is a promising approach to empirical forecasting with such datasets.

\section{Overview of the Methodology}\label{sec:Overview} 

A probabilistic forecast \citep{Gneiting_etal_2014} over a continuous random
variable of interest $Y$
given observations governed by random variables $X$ is a
conditional distribution $P(Y|X)$ with density $p(y|x)$, that is, a mapping from a predictor $x$
bearing information about $Y$ to a distribution over possible
outcomes of $Y$. It is well-understood that the essence of the probabilistic forecast is
to provide the ability to fully describe the uncertainties in predicted outcomes.

In the case of probabilistic forecasts of weather, $x$ can be made up of weather
conditions at various spatio-temporal locations, such as might be provided by
readings from an array of weather stations. Alternatively, $x$ might comprise
output data from an ensemble of numerical weather prediction (NWP) simulations,
which in turn are functions of weather observations through data assimilation
procedures. The first case corresponds to what are termed ``empirical''
forecasts \citep{van2007empirical}, whereas the latter case corresponds to
``physics-based forecasts'' \citep{toth2019weather}. Our purpose in this work is
to focus on the information-theoretic aspects of machine learning-mediated
forecasts, and how information embedded in predictors concerning predicted
quantities may be maximized. From this perspective, the distinction between the
two types of predictors is largely immaterial.

It is useful to introduce the idea of the unique \textit{ideal forecast
distribution} \citep{gneiting2013combining,Gneiting_etal_2014,graziani2019} with
density $p(y|x)$, which makes maximal use of the information in $x$. It is
intuitively straightforward to see that such a unique distribution exists, as it
would be empirically ascertainable in principle by first estimating the joint
density $p(x,y)$ by histogramming a large dataset of pairs $\{(x_l,y_l):l=1,\ldots,N\}$
($N\gg 1$), and conditioning to estimate $p(y|x)$. A true first-principles ideal
forecast distribution for weather is essentially unobtainable, since it would
require modeling precision and computational power far beyond what is currently
available \citep{fritsch2004improving,knutti2013robustness}. However as the
thought experiment indicates, data-driven approximations to ideal forecast
distributions may be obtained given sufficient data. 

Unfortunately, it is usually the case that the data $x$ is unmanageably large,
consisting not only of many samples (which is desirable) but of very large
vectors (which is less so). To make progress, $x$ must be processed by a
transformation $T(x)$ that effects a dimensional reduction. Clearly, $T(x)$ must
be chosen with a view to limiting damage to the information that $x$ bears with
respect to $y$. As we will see below, it is often quite practical to achieve
this, because there is frequently much more data mass in $x$ than is strictly
necessary to code information concerning $y$. Making judgments concerning the
appropriate choice of reductions $T(\cdot)$ requires comparing the properties of
the distributions $P(Y|X)$ and $P(Y|T(X))$, as we will discuss below.


Another concept that we make essential use of below is that of probabilistic calibration
\citep{gneiting2013combining,Gneiting_etal_2014}. Probabilistic calibration is a
desirable feature in a forecast because it means that the forecast is ``honest''
about the probabilities of its quantiles, since those probabilities correspond
to long-term average frequencies that are expected in a forecast system based on
the reduction $T(\cdot)$ that furnishes an approximation to the ideal forecast distribution
$P(Y|T(X))$. 

As mentioned above, we approach the probabilistic forecasting task in two main steps. The first is
learning of a suitable $T(\cdot)$ that reduces the dimension of predictors
$x$ while retaining as much information about the response as possible. The
second is joint modeling of the distribution of the artificial predictors
$T(x)$ and the response. Once the joint modeling is complete, a numerical
integration of the modeled density yields the forecast, i.e. the density
$q(y|T(x))$ corresponding to the distribution $P(Y|T(X))$.

Reduction of dimension via $T(\cdot)$ is necessary, because the number of data
samples required to model a joint density essentially increases exponentially in
the number of dimensions to be considered. Despite this, the goal is
to make forecasts as precise as possible, while still retaining calibration.
This is the principle of ``maximizing sharpness subject to calibration''
\citep{gneiting2007probabilistic}. Thus, a major part of this work is the
efficacy of the dimension reduction technique.

In our experience with our two-step forecasting procedure, the problem of maximizing sharpness subject to calibration 
has turned out to separate into two essentially independent problems.
In the first step, we maximize forecast ``sharpness'' --- a measure of concentration of the
forecast distribution to small outcome sets \citep{gneiting2007probabilistic}
--- by minimizing the Kullback-Leibler divergence between $P(Y|X)$ and
$P(Y|T(X))$, which turns out to be equivalent to maximizing the mutual
information between $Y$ and $T(X)$ with respect to the choice of $T()$. This is described in 
\S\ref{sec:DR}. We verify the increased sharpness of our computed
forecasts relative to that produced using plausible alternative dimensional
reductions using entropy games \citep{graziani2019}, a recasting of ``ignorance
scoring'' \citep{Roulston_Smith-2002}, as described in \S\ref{sec:PPF}.

Calibration then comes down to properly modeling the joint distribution of the 
low dimensional representations $T(X)$ and the responses $Y$.
We test calibration by observing what percentage of test set examples
fall in modeled probability contours in our forecasts. Specifically, for each
predictor in the test set, we can forecast the response using our models. Each
forecast is a density of response, and so we can compute probability contours
for this density. Over the entire test set, we can see what percentage of true
test responses fall within the modeled $.683$ and $.954$ contours. Of course, if
the forecasts are in line with the data generating distribution, then up to
noise in sampling the test set from the data generating distribution and rounding errors, we should
see $.683$ and $.954$ of the test samples fall within the respective
modeled contours. Though this is obviously not a sufficient condition for
calibration, it is a useful necessary condition that gives rough idea of how
honestly our forecasts deal with uncertainty. This procedure is described in
greater detail in \S\ref{sec:PPF}. 

\section{Description of the Data, their Handling}\label{sec:Data}
In this section, we describe the datasets used to verify our machine learning methodology. This description of the data is useful to have in advance of
the detailed methodological development, since it can help clarify and motivate the
technical aspects of our work.


The data are zonal and meridional winds that are the output of historical simulations based on a widely used mesoscale weather model, Weather and Research Forecasting (WRF) model; the simulations were run between the years 1980-2010 at a grid resolution of 12 km x 12km and with output saved once every 3 hours \citep{wang2014downscaling}. 
 We consider in particular a 19x19 box of spatial locations around Kansas City at highest resolution afforded by the simulation, between the years 1984 and 2005. We specify the task of forecasting
these two components of wind at $t+3$ hours 
at the center of our grid, using all wind data from
around the grid at times $t, \hdots, t-k$.  The 3
hour forecasting window is specified so that measurements from such a relatively 
small box of values are informative of the response; forecasting weather further into the
future would require a larger grid of observations. 

We formulate examples $(y_i,
x_i)$ by letting $y_i$ be the the 2-dimensional  wind
vector 
at the center of the grid at  timestep $i+1$, and letting $x_i$
be the concatenation of wind vectors around the grid at time steps $i, \hdots,
i-k$, for some positive integer $k$. Given the small size of the box and the characteristic time- and length scales of the wind patterns,
not much information about the response is contained when $k >3$, so we settle
on $k=3$. 

For any given model that we train, we restrain the model to consider small
``seasonal'' slices of three months of each year. Thus, a typical model is
trained using some number of years of ``winter'' examples, and likewise
validated on winter examples; by ``winter'' we specifically mean the 3 months of 
year's worth of 3-hour timesteps, by ``spring'', the second
three months, etc. 
 We treat $(y_i, x_i)$ as though they are drawn independently and
identically distributed from some distribution $\mathcal{D}$; the restriction to
seasonal slices simplifies the process of approximating $\mathcal{D}$
significantly. Secular drift of the weather is a concern, but in
training we avoided this issue by ignoring the temporal positions of
the samples. 

Our training, validation, and testing procedures are as follows. We first separate a test set of the
observations from the most recent seasonal slice, e.g. the ``winter'' from the
most recent year in a specified range of years. We then take winter observations
from the previous 10 years, and divide these examples uniformly at random
into dimension reduction training, joint model training, and validation sets.
The validation set is used for model selection in both the dimension reduction
and joint modeling. The test set is not consulted until after all final
modeling decisions have been made. We observe the test set to measure forecast
precision and confirm calibration of our forecasts on true holdout data, and in
particular, holdout data from a never before seen season in the ``future''. Here
secular drift of the weather does come into play, and so confirmation of
calibration on these future weather samples is vital.

\section{Dimension Reduction}\label{sec:DR} 
Dimension reduction of the predictors is a necessary step in any continuous
probabilistic forecasting scheme based on a dataset with high dimensional features,
since density estimation in high-dimensional spaces is well-known to be
challenging \citep{silverman2018density}. Naturally, this is often the case for geoscientific data, where large spatial and temporal variabilities abound. To understand the most important patterns (including time and space) of a dataset, one needs to use dimensional reduction technique to reduce the data to certain dimension, from which one can understand the importance of each pattern (in space) and how that pattern change with time. However, fully unsupervised dimension reduction techniques such as Principal Component Analysis (PCA) are inadequate for the purposes of forecasting, as they only project predictors onto informative subspaces, ignoring the relationship of predictors and response. With this in mind, we employ a dimension reduction technique that is focused on preserving as much
information about the response as possible. 

\subsection{Information Preserving Reductions}\label{subsec:IPR}
Information theory furnishes a useful framework in which to explore these ideas formally, and is at the basis of the method we use. Indeed, 
there is a well-developed literature of dimension reduction for supervised learning that follows the same ideas as 
we use here; see for example \citep{murillo2007MI} and \citep{shadvar2012MI}.

We briefly review the required
information-theory background \citep[see][for a detailed treatment]{thomas1991information}.
Given a pair of random variables $(Y,X)$ with a joint distribution $P(X,Y)$
assumed absolutely continuous for simplicity, and hence possessed of a density
$p(x,y)$; and given marginal distributions $P(X)$, $P(Y)$ with densities $p_x(x)$,
$p_y(y)$, the mutual information $I(Y;X)$ is defined by 
\begin{equation}
    I(Y;X) = KL\left[p(x,y) || p_x(x)p_y(y)\right], 
\end{equation}
where $KL$ denotes the Kullback-Leibler divergence, given by the formula
\begin{equation}
KL\left[p(z) || q(z)\right] \equiv \int dz\,p(z)\log\left[\frac{p(z)}{q(z)}\right].
\end{equation}
The Kullback-Leibler divergence \citep{kullback_leibler_1951} is a measure of
the difference of two distributions. It has the property of being bounded below
by zero, a value attained when the densities $p(z)$ and $q(z)$ are equal almost
everywhere. It follows from this the mutual information $I(Y;X)$ is small when
$p(x,y)$ can approximately be factored into marginals, i.e. when $X$ hardly any
information about $Y$. 

Now consider a dimensional reduction $T: \mathbb{R}^d \to \mathbb{R}^m$, $m<d$.
The forecast distribution conditioned on $X$ is $P(Y|X)$. As we discussed in
\S\ref{sec:Overview}, this distribution is in principle (if perhaps not in
practice) empirically ascertainable by histogramming a large database of pairs
$(y,x)$. On the other hand, after reduction by $T(\cdot)$, we are concerned with
the forecast distribution $P(Y|T(X))$. This distribution is in principle
``degraded'' by information loss with respect to $P(Y|X)$.  The KL divergence and
the mutual information are tools that allow us to choose $T()$ such that these 
two distributions are as close as possible given $m$, up to sampling and optimization considerations.

The reduction $T(\cdot)$ implicitly defines a new random variable $\hat{T}\equiv T(X)$.
We proceed by considering the KL divergence of $P(Y|X)$ and $P(Y|\hat{T})$,
\begin{eqnarray}
K(X)&\equiv&
KL\left[P(Y|X) || P(Y|\hat{T})\right]\nonumber\\
&=&\int dy\, p(y|X)\log\left[\frac{p(y|X)}{q(y|T(X))}\right].
\label{eq:kl_redux}
\end{eqnarray}
Since we have a sample of realizations from the random variable $X$ from which
to estimate this quantity, we average Equation~(\ref{eq:kl_redux}) by taking the
expectation under $P(X)$ using the corresponding density $p(x)$:
\begin{eqnarray}
\mathcal{E}_{X\sim P(X)}\left\{ K(X)\right\}
&=&\int dxdy\,p(x)p(y|x)\nonumber\\
&&\quad\times\log\Bigg[\frac{p(y|x)}{q(y|T(x))}\Bigg]\nonumber\\
&=&\int dxdy\,p(x,y)\nonumber\\
&&\quad\times\log\Bigg[\frac{p(x,y)q(T(x))}{p(x)q(T(x),y)}\Bigg].
\label{eq:mi_1}
\end{eqnarray}
In Equation~(\ref{eq:mi_1}) we have introduced new densities $q(t)$ and $q(t,y)$
corresponding to the distributions $P(\hat{T})$, $P(\hat{T},Y)$, where $t$ is a
realization of $\hat{T}$. We may now insert a factor of $p(y)$ into the numerator and denominator
of the log term in
Equation~(\ref{eq:mi_1}), and split the log
term, to obtain
\begin{equation}
\mathcal{E}_{X\sim P(X)}\left\{ K(X)\right\}=
I(Y;X)-I(Y;\hat{T}).
\label{eq:mi_2}
\end{equation}

In Equation~(\ref{eq:mi_2}), only the second term depends on the choice of the
reduction function $T(\cdot)$. We can therefore see that choosing a reduction
that minimizes the average information divergence between $P(Y|X)$ and
$P(Y|\hat{T})$ is equivalent to choosing
$T(\cdot)$ to maximize the mutual information between $Y$ and $\hat{T}=T(X)$.

The approach we take it is train a neural network $T: \mathbb{R}^d \to
\mathbb{R}^m$ from a shallow but flexible class of networks by maximizing an
empirical estimate of the expected mutual information between $Y$ and $T(X)$
with respect to the parameters defining the network. In the next subsection we
describe practical implementations of this approach.

\subsection{Computing and Maximizing the Objective}

In principle, the goal of optimizing the dimensional reduction $T(\cdot)$ can be
attained by choosing as maximization
objective an empirical estimate for the average KL-divergence:
\begin{equation}
\max_{T \in \Theta} \  \frac{1}{N}  \sum_{i=1}^N \log\bigg(\frac{q\big(y_i, T(x_i) \big)}{q\big(T(x_i)\big)} \bigg), 
\label{eq:objective}
\end{equation}
where $\Theta$ denotes the set of all functions that can be represented by
whatever neural network architecture chosen. There is a difficulty, however, in
that estimates of the densities $q(y,t)$ and $q(t)$ must be produced at each
iteration, because the densities depend on the transformation $T(\cdot)$, and so
evolve as $T(\cdot)$ evolves. This circumstance compounds the computational challenge already
inherent in empirical density estimation.

We experimented with $k$-nearest neighbors density estimation. Essentially, once $k$ has been specified, the density estimate for any given point $(y_i, T(x_i))$ is
\begin{equation*}
\hat{\pi}\big(y_i, T(x_i)\big) = \frac{k}{N} \frac{1}{Vol_k\big(y_i, T(x_i)\big)},
\end{equation*}
where $Vol_k(y_i, T(x_i))$ is the volume of the $k^{th}$ smallest parallelepiped
in the set of $p+m$ dimensional parallelepipeds defined by $(y_i, T(x_i))$ and
the other points in the training set. In order to get good density estimates,
large minibatches are required in the optimization, which slows down convergence
to an unacceptable degree. In the end, we opted to abandon this approach due to
fragility and poor performance.

Another option for estimating the densities in Equation~(\ref{eq:objective})
would be to resort to the probabilistic machine learning technique of
normalizing flows that we describe in \S\ref{sec:PPF} below, which we use to
estimate the forecast density. This would have the effect of inserting a second
expensive ML training loop in our workflow, and probably turning the problem
into a true high-performance computing (HPC) enterprise. This approach is worth
considering in the future, but for simplicity in this work we adopted a 
normal-theory approximation that we now describe. 

The objective is radically simplified if we simply approximate the joint
distribution of $(Y,\hat{T})$ as Gaussian. Suppose this assumed normal
distribution is governed by a covariance given in block form by
\begin{equation}
\Sigma\equiv\left[
\begin{array}{cc}
\Sigma_{YY}&\Sigma_{T\hat{T}}\\
\Sigma_{\hat{T}Y}&\Sigma_{\hat{T}\hat{T}}.\\
\end{array}
\right]
\label{eq:covariance}
\end{equation}
Then using standard normal theory arguments one can show that the mutual
information $I(Y;\hat{T})$ is given by the expression
\begin{eqnarray}
I(Y;\hat{T})&=&-\frac{1}{2}\ln\left[\det
\left(\Sigma_{YY}-\Sigma_{Y\hat{T}}\Sigma_{\hat{T}\hat{T}}^{-1}\Sigma_{\hat{T}Y}\right)\right]
\nonumber\\
&&+\frac{1}{2}\ln\left[\det\Sigma_{YY}\right].
\label{eq:gaus_approx_1}
\end{eqnarray}

Thus in this approximation, minimization of the expected KL divergence in
Equation~(\ref{eq:mi_2}), which is equivalent to maximization of $I(Y;\hat{T})$,
is in turn equivalent to
\begin{equation}
\min_{T \in \Theta} \ 
\ln\left[\det
\left(\Sigma_{YY}-\Sigma_{Y\hat{T}}\Sigma_{\hat{T}\hat{T}}^{-1}\Sigma_{\hat{T}Y}\right)\right],
\end{equation}
that is, to minimizing the predictive variance of $y | T(x)$. 

We implement this approximation by computing empirical data covariances from the
data to obtain the elements of the array in Expression~(\ref{eq:covariance}). We
have found the resulting data reduction procedure to be far more rapid and
robust that any based on density estimation, despite the occasionally
questionable nature of the Gaussian assumption for data that clearly exhibits
non-Gaussian features.

\subsection{Target Dimensions and Sufficient Dimension Reduction}

It is a known result of normal theory that given that a response $y$ of
dimension $p$ and predictors $x$ of dimension $d$ that have a jointly normal
distribution in $p+d$ dimensions, there exists an affine transformation $T:
\mathbb{R}^d \to \mathbb{R}^p$ that is fully information preserving. More
precisely, there exists an affine $T: \mathbb{R}^d \to \mathbb{R}^p$ such that 
\begin{equation}
 \forall x,  KL\left[n_x(y|x) || n_t(y|T(x))\right] = 0,
\end{equation}
where $n_x(y|x)$ is the normal predictive density of $y$ conditioned on $x$, and
$n_t(y|T(x))$ is the normal predictive density of $y$ conditioned on $T(x)$.

This result has an intuitively reasonable explanation: a normal distribution is
specified by its mean and variance alone, and since the conditional variance of
$y|x$ is independent of $x$, then given the joint covariance structure of
$(y,x)$, the family of distributions $y|x$ is entirely specified by the
conditional mean, which is a $p$-dimensional linear function of $x$. The
conditional mean is in fact the dimensional reduction $T(x)$ in this case. In
the statistics literature, the idea of the distribution of the response
conditional on the predictors being invariant to a specific transformation of
the predictors into some lower dimensional space is known as ``sufficient
dimension reduction," \citep{adragni2009sdr}.

\begin{figure}[t]
\centering
\includegraphics[width=.35\textwidth]{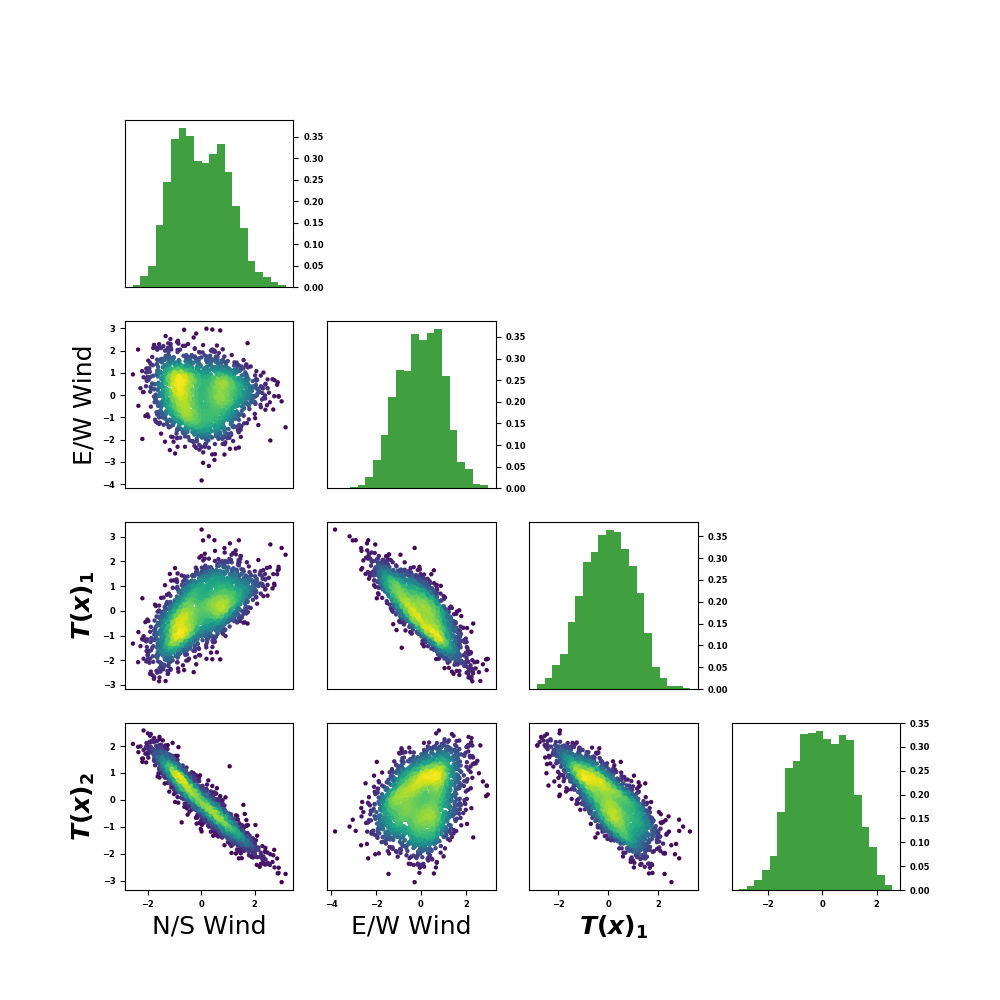}\hfill
\includegraphics[width=.35\textwidth]{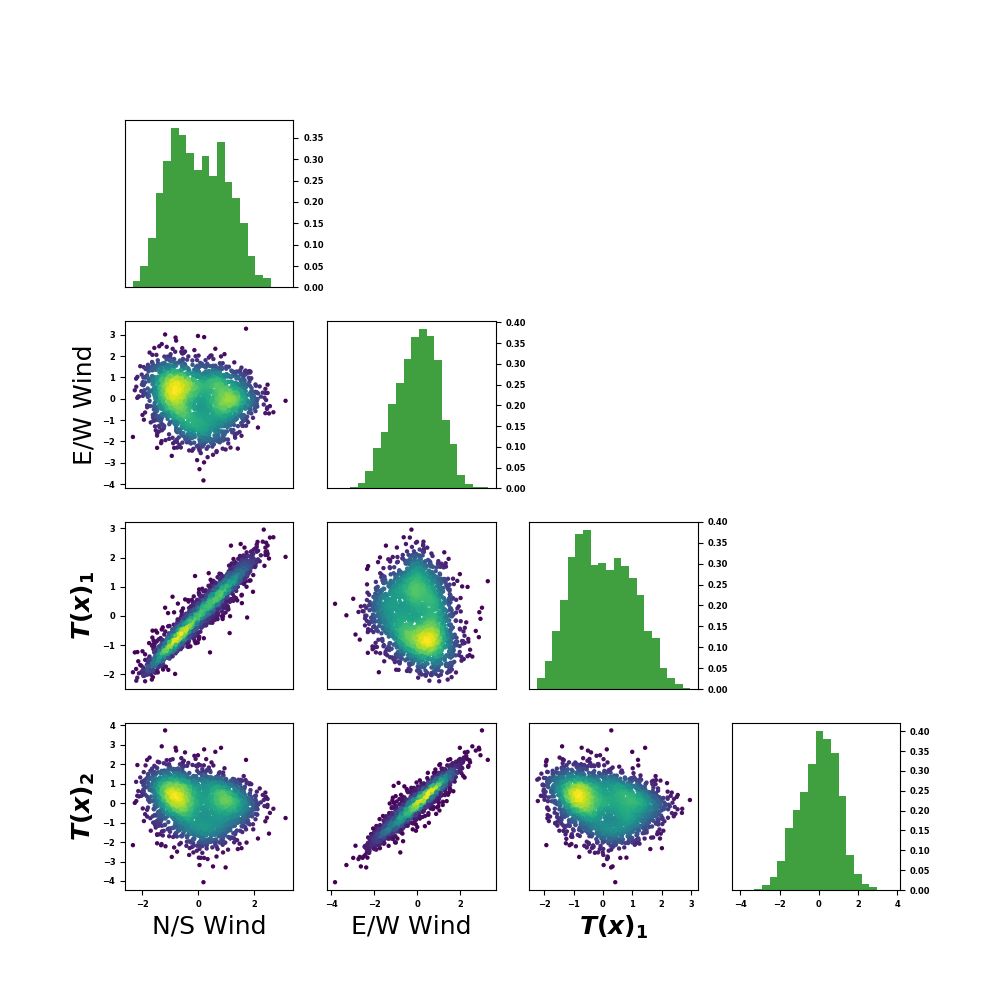}\hfill
\includegraphics[width=.35\textwidth]{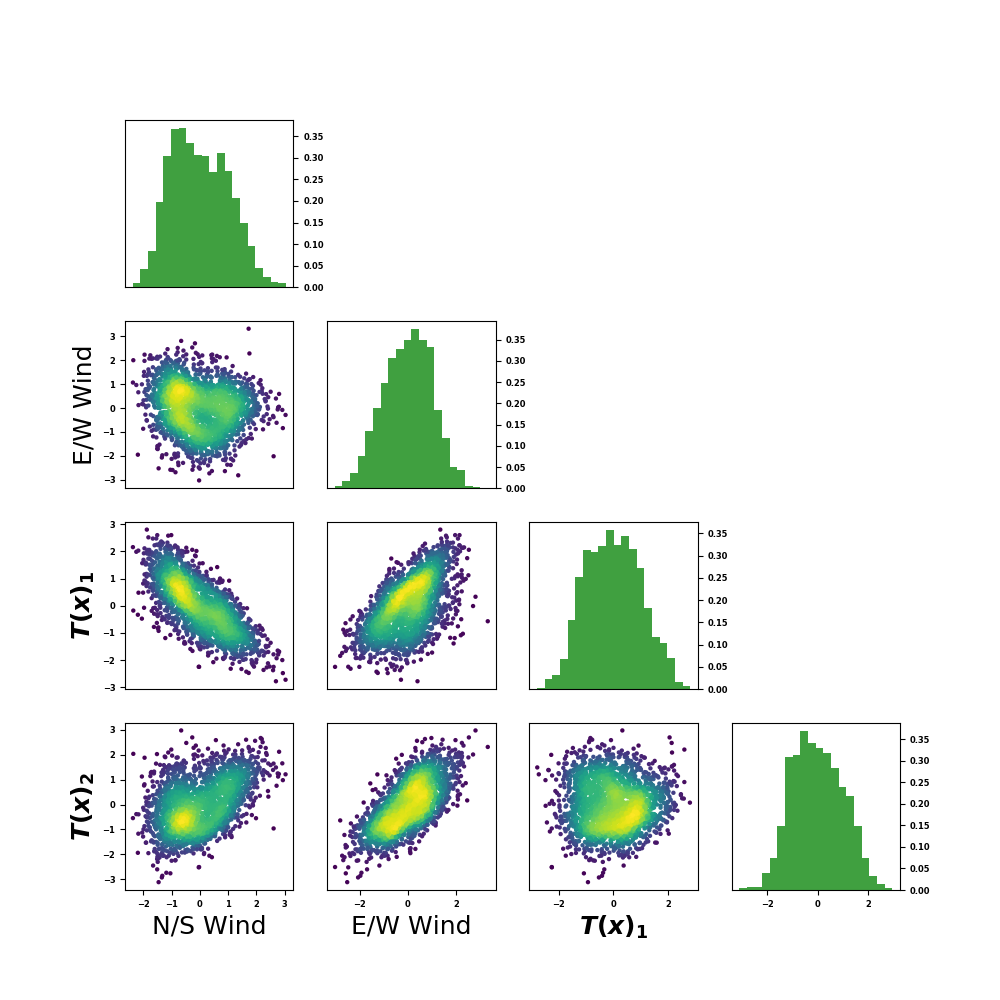}

\caption{Plots of the training data resulting from the three dimension reduction techniques used in the experiments on the Jan-March seasonal slice, with Year 1 response. On the top, the result of the information preserving reduction with normal approximation, in the middle, grid point choosing, and finally PCA at the bottom. $T(x)_1$ and $T(x)_2$ denote the artificial predictors; the map $T$ is learned in dimension reduction.}
\label{fig:reduced_data}
\end{figure}

The relevance for our task is that that the reduced dimension $m$ should be
expected to be commensurate --- if not precisely equal, as in the normal theory
case --- to the output dimension $p$, so that when the output dimension is much
smaller than the input dimension of the predictors, radical dimensional
reductions are in principle possible. In the case of our simulated wind data,
where we predict the two wind components at one grid point, we would expect a
very considerably reduced dimensionality ($m=2$, in fact) in the normal case.
For many applications where data are approximately jointly normal, there's no
reason to believe that $m \gg p$ reduced predictors are required to capture the
majority of the information stored in the $d$ dimensional raw
predictors.\footnote{Note that the reductions that we consider here are only
information-preserving \textit{with respect to the chosen forecast quantities},
and not preserving of general information. They are in general expected to be very
information-lossy with respect to other predicted quantities.} 

In practice, we have indeed set $m$ to be $p$ in our numerical experiments. Of
course, higher dimensional $m$ could and should be explored, depending on the
size of the dataset at hand. For large data sets, it is possible that the extra
dimensions for artificial predictors are helpful in sharpening
forecasts at moderate extra cost, but this may not be the case for smaller
datasets, where a lack of training examples in a higher dimensional regime may
make calibration of the forecasts hard or impossible.

Despite the fact that our data are by no means jointly normal 
this normal
approximation has proved to be effective in producing sharp forecasts and lower
training times. There is some approximate normality in the raw predictors, which
may be part of the reason we have found small $m$ sufficiently information
preserving.

\section{Producing Probabilistic Forecasts}\label{sec:PPF}

Given a dimension reducing transformation $T(\cdot)$, the forecasting task next
demands a joint modeling of the response $y_i$ and the artificial predictors
$T(x_i)$. Given the an estimate for the joint distribution of response and
predictors, forecasts can be produced by standard numerical integration
techniques, as probabilistic forecasts are simply conditional distributions.

\subsection{Normalizing Flows}
We model the joint distribution by a neural network-based non-parametric density
estimation technique referred to in the literature as the ``normalizing flow''
\citep{kobyzev2019}. We briefly review normalizing flows here, consigning more
detail to Appendix \ref{appendix:NF} 

Given a random variable $W$ taking values in $\mathbb{R}^{p+m}$ with density
function $p_{W}$, recall that the formula for the density of the random variable
$V$ formed by mapping $V$ through the differentiable bijection $\phi:
\mathbb{R}^{p+m} \to \mathbb{R}^{p+m}$; $W=\phi(V)$ is
\begin{equation} 
p_V(v) = p_W\big(\phi(v)\big) \cdot |\partial \phi |,
\label{eq:diffeo}
\end{equation} 
where $\partial \phi$ denotes the Jacobian matrix of the transformation. 

The idea of normalizing flows is thus to fix some $W$ as the ``latent'' random
variable with a known simple distribution density (often a standard normal
density or a normal mixture), and given samples of some random variable $V$
taking values in $\mathbb{R}^{p+m}$, and with with unknown density $p_V(v)$,
model $p_V$ by learning a differentiable bijection $\phi$ that satisfies
Equation~(\ref{eq:diffeo}). Assume that the latent distribution $p_W$ can be
parameterized by $\omega \in \Omega$. We express $\phi$ as a neural network in
some flexible class of functions parameterizable as neural nets $\Phi$, and then
simply learn $\phi$ by maximizing the log-likelihood of the data in the latent
space, i.e. we consider the optimization objective
\begin{equation*}
\max_{\phi \in \Phi, \ \omega \in \Omega} \  \frac{1}{N} \sum_{i=1}^N \log\bigg( p_W\big(\phi(v_i); \omega\big) \bigg) + \log\bigg( |\partial\phi(v_i)|\bigg),
\end{equation*}
and use a gradient-descent approach simultaneously on $\omega$, and the parameters characterizing $\phi$.

\subsection{Training Joint Distributions for Calibration}

One of the canonical issues in machine learning practice is overfitting
\citep{goodfellow2016deep}. In a
nutshell, overfitting is the phenomenon of an extremely expressive model
beginning to model ``features'' in the training data that are due to sampling
noise, rather than to the generating distribution. The result is well-understood
to be a degradation in the generalization capacity of the model. In the context
of forecasting, this is tantamount to loss of forecasting skill.

\begin{figure}[t]
\centering
\includegraphics[width=.5\textwidth]{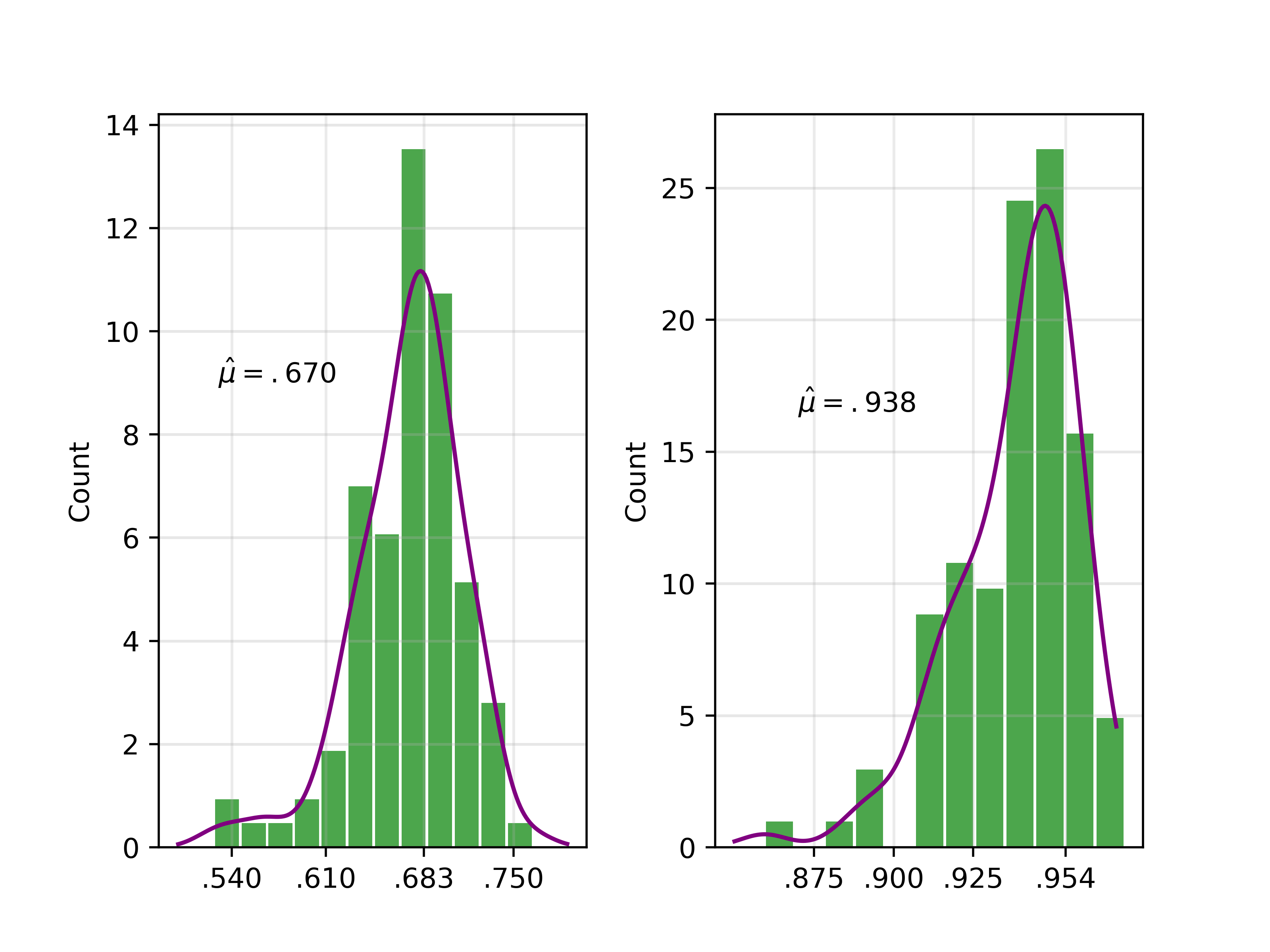}\hfill
\caption{A histogram constructed from the test set hit rates generated on each
year/season combination. The fact that these histograms are approximately centered
at .683 and .954 without too much variance is strong evidence that our forecasts are indeed 
calibrated. We do see that the average hit rates are in both cases a bit too small, meaning that 
our forecasts tend to be slightly ``too certain'' about the future.}
\label{fig:calibration_results}
\end{figure}

Regularization is the general term for any strategy that combats overfitting.
One type of regularization is achieved by monitoring performance on a validation
set, and picking the model that performs best on this validation set. To train
for calibration, we retain this philosophy, although perhaps not in the most
standard way. 

Instead of picking the model with best validation loss, we employ an early
stopping strategy centered around achieving calibration on the validation set. 
For a probabilistic forecast to be useful at all it is necessary that it be
probabilistically well-calibrated, and we therefore design our regularization
strategy around calibration. We train the model for some
relatively large number of iterations, well beyond 
the point where the training loss is no longer obviously improving; we stuck to 1200 
steps of gradient descent for our simulated wind data experiments, 
but learning rates, etc will play an important role in how many iterations this takes. 
Every 100 training steps after some initial learning phase where our joint model is 
clearly still experiencing drastic improvement on the training set, we consider the current learned model, and do a numerical
integration to produce a forecasting machine. With this interim forecasting machine, 
we can check calibration on the validation set, and in the end, we chose as our final
forecasting machine the interim model that was best calibrated on the validation set. 

The notion of ``best calibrated'' was somewhat arbitrarily chosen. As mentioned above, each time 
we check calibration, we compute the percentage of validation responses
falling within the modeled $.683$ contour, and $.964$ contour.  To compute a calibration score 
$s_c$ for a given model, we calculate the absolute deviations from nominal contour calibration, and we use the following convex combination of those deviations:
\begin{equation*}
s_c := \frac{13}{23} | .683 - hr_{.683} | + \frac{10}{23} | .954 - hr_{.954} |, 
\end{equation*}
where $hr_p$ is the computed percentage of validation examples falling 
in the model contours, what we often refer to as the ``hit-rate" of the $p$ contour. The slight 
weighting towards the inner contour reflects the observation that often inner contours 
were more difficult to model in our particular case, as much of the joint distribution
detail tends to happen near modes. From a larger methodological perspective,
this should simply be viewed as another hyperparameter. 

\section{Experiments}\label{sec:Exp}

Our numerical experiments to validate the forecasting software were conducted
using the simulated wind data described in \S\ref{sec:Data}.

\begin{figure*}[t]
\centerline{\includegraphics[width=\textwidth]{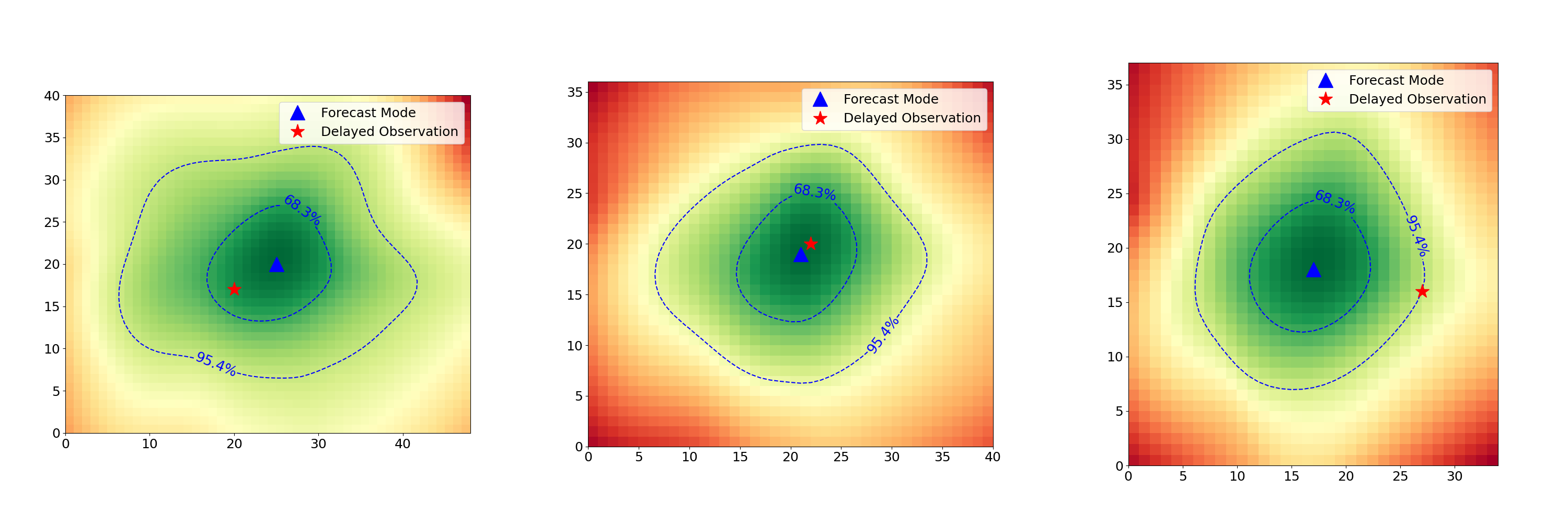}}

\caption{
Visualizations of example
forecasts, providing an impression  of the hit rate
computations. Hit rate computations are performed by computing what percentage of delayed observations fall in the modeled contours.
} \label{fig1}
\end{figure*}

A given experiment is carried out as follows. We fix a recent year in the range
1995 to 2003, and a season in that year. The year
and season choice specify the test set, and the training and validation sets are
then constructed by taking the examples appearing in the 10 previous examples of
the chosen season. We then produce forecasts, and report performance on the test
set. The use of 10 previous years is arbitrary; in practice, this figure
should be treated as a hyperparameter governing the degree to which the model is
informed by recent events. 

Performance is measured in two ways. Firstly, we test the models for calibration by 
computing the ``hit rates'' for the .683 and .954 contours on test data, just as 
was done in evaluating calibration of interim models on the validation data 
during training. We do this for each experiment's test set, yielding a set of hit rate scores
on hold out data. We give histograms of these empirical measure of calibration over this set of scores
in Figure \ref{fig:calibration_results}, giving evidence of approximate calibration via our 
methods. This is vital, as if forecasts are not properly calibrated, they can be unrealistically 
certain about the events they forecast, and essentially cheat in competitions of 
forecast precision, whereby sharper forecasts are preferred, but only conditional 
on calibration.

After guaranteeing that calibration has been approximately achieved, we can
begin to consider precision, or ``sharpness'', of forecasts. Assuming proper
joint modeling of response and predictors, the sharpness of the forecasts is a
direct result of the efficacy of the dimension reduction; it tells us how much
information about the response is being ``saved'' in our low dimensional
representation. As such, we use this section to display the advantages of our
dimension reduction technique, even with the crude approximation of joint
normality in lower dimensions. 

As mentioned above, one might naively consider picking $m$ locales that are
highly correlated with the response as a form of crude dimension reduction. This
is actually a relatively reasonable approach, as some variables at some points
on the grid have correlation >.95 with corresponding response variables 3 hours
in the future. Another dimensional reduction technique that is familiar to
machine learning practitioners is PCA. There is
no reason to think that PCA is likely to furnish informative data reductions in
the current setting, as we are concerned with safeguarding predictor information
concerning responses about relationships between response and predictor, and PCA
is an unsupervised algorithm that when applied to predictors is agnostic about
responses. Nonetheless we include it, given its popularity, as a sort of
baseline reduction.

To test sharpness, we play a information theoretic game called the Entropy Game
\citep{graziani2019}, a re-elaboration of the Ignorance Score
\citep{Roulston_Smith-2002}, between forecasts. This amounts to evaluating the
log-likelihood of the conditional distribution corresponding to each forecast
for each instance in the test set at the observed value, and computing the difference in the sums of
these log-likelihoods over the test set. Basically, a forecast does well in the
entropy game if the evaluated log-likelihoods are usually large at the wind
values that were actually observed, which given
that probability density functions are normalized, is a reasonable measure of
how sharp the forecasts are. As discussed in \cite{Roulston_Smith-2002} and \cite{graziani2019},
a forecast that exhibits superior sharpness in this sense can furnish the basis
for winning wagers on forecast outcomes against less sharp forecasts, and hence
is in a realizable sense a basis for superior decision-making.

\begin{figure}[t]
\centering
\includegraphics[width=.45\textwidth]{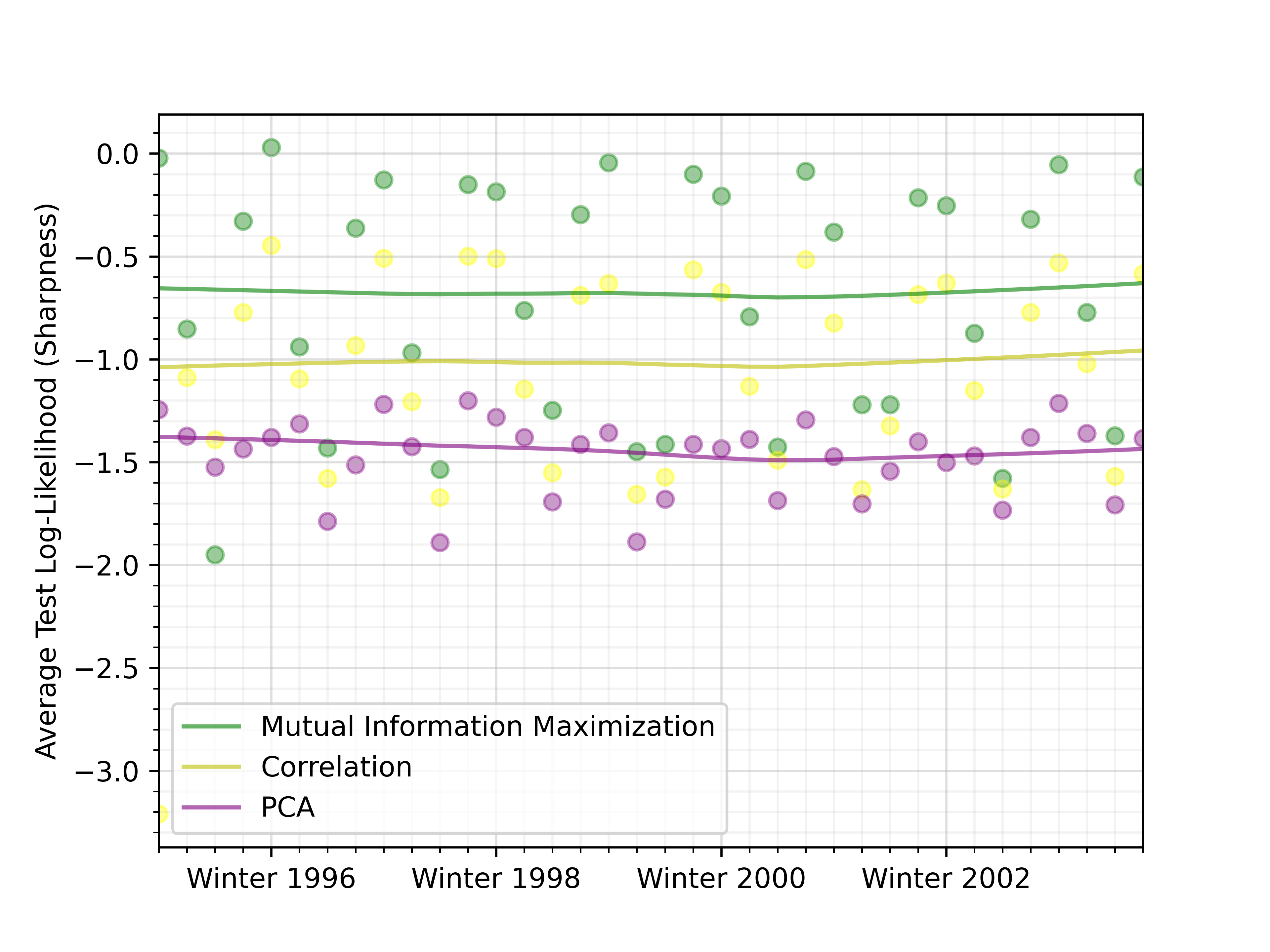}\hfill
\caption{A visualization of the test likelihoods for each of the three dimension reduction techniques employed, for each of the year/season experiments run. It is very clear that the performance of the information theoretic based dimension reduction technique is an improvement over more naive methods.}
\label{fig:sharpeness}
\end{figure}

Figure \ref{fig:sharpeness} shows the sharpnesses of forecasts reduced by the
three methods. It is evident from the figure that the normal theory-based
dimension reduction has improved sharpness over naive grid point picking, wherein the most correlated locales with future observations are used as 
predictors; this baseline outperforms PCA, as expected.

\section{Discussion}\label{sec:Discussion}

The fact that calibration can approximately be achieved is not necessarily
obvious \textit{a priori}. After all, in making up the test set with future
observations, we are directly subjecting ourselves to confounding effects of
secular weather drift. That being said, calibration on the test set is really a
statement about the invariance of the modeled conditional distributions, and it
seems clear that  these conditional distributions should be much less affected
by secular drift than the marginals over the predictors we condition on, i.e.
the climatetology. This seems to be a key to the viability of this proposed
procedure.

A potential drawback to our approach is the fact that we did not settle
on the final experimental design until after some exploratory
experimentation with our models on the data set. It is therefore
possible that we implicitly learned something about the hyperparameter
settings of our models while analyzing different training-test set
splits. We intuitively expect (but have not explicitly shown) that
out-of-sample generalization failures due to such model dependencies are
weak compared to well-known failures due to in-sample parameter overfitting.

In fact, it is very possible that better results are achievable. In the experiments
reported here, we did not exhaustively search hyperparameter space in even in
the most important dimensions. For example, the frequency of validation set calibration
checks, the weighting towards inner contour validation estimates, the number of
dimensions we reduce to, the search over novel architectures in the learning of
dimension reductions, and the number of previous seasons to add into the
training/validation set are all candidates for  investigations that
could improve performance. As mentioned above, the choice of updating the model
seasonally is arbitrary. It's quite possible that more frequent updating could
improve performance. 

As noted in \S\ref{sec:DR}, it would be possible to improve on normal theory
reduction by incorporating a second normalizing flow training loop at the
dimensional reduction stage, so as to obtain estimates of the probability
densities in the expression for objective function of
Equation~(\ref{eq:objective}). This training would have to be repeated at every
step of the gradient descent over the parameters of the reduction function
$T(\cdot)$. This would in principle make the computation a truly HPC-scale
problem, although it is possible to imagine ways of abating the cost by making
the two optimizations --- over $T(\cdot)$ and over NF parameters --- aware of
each other.  The benefit of such an approach would depend on the degree to which
the non-Gaussian nature of the data limits the information-preservation
properties of the normal theory approximation.

\section{Acknowledgements}
This material is based upon work supported by Laboratory Directed
Research and Development (LDRD) funding from Argonne National
Laboratory, provided by the Director, Office of Science, of the U.S.
Department of Energy under Contract No. DEAC02-06CH11357.

\section{Data Availability Statement}
All data and code used in this work are publicly accessible via the following repository: \url{https://github.com/rittlern/probabilistic_forecasting}.

\section{Government License}
The submitted manuscript has been created by UChicago Argonne, LLC,
Operator of Argonne National Laboratory (``Argonne''). Argonne, a U.S.
Department of Energy Office of Science laboratory, is operated under
Contract No. DE-AC02-06CH11357. The U.S. Government retains for itself,
and others acting on its behalf, a paid-up nonexclusive, irrevocable
worldwide license in said article to reproduce, prepare derivative
works, distribute copies to the public, and perform publicly and display
publicly, by or on behalf of the Government. The Department of Energy
will provide public access to these results of federally sponsored
research in accordance with the DOE Public Access Plan \url{https://www.energy.gov/downloads/doe-public-access-plan}.


\bibliographystyle{ametsocV6}
\bibliography{paper}


\appendix

\begin{figure*}[h]
\centerline{\includegraphics[width=\textwidth]{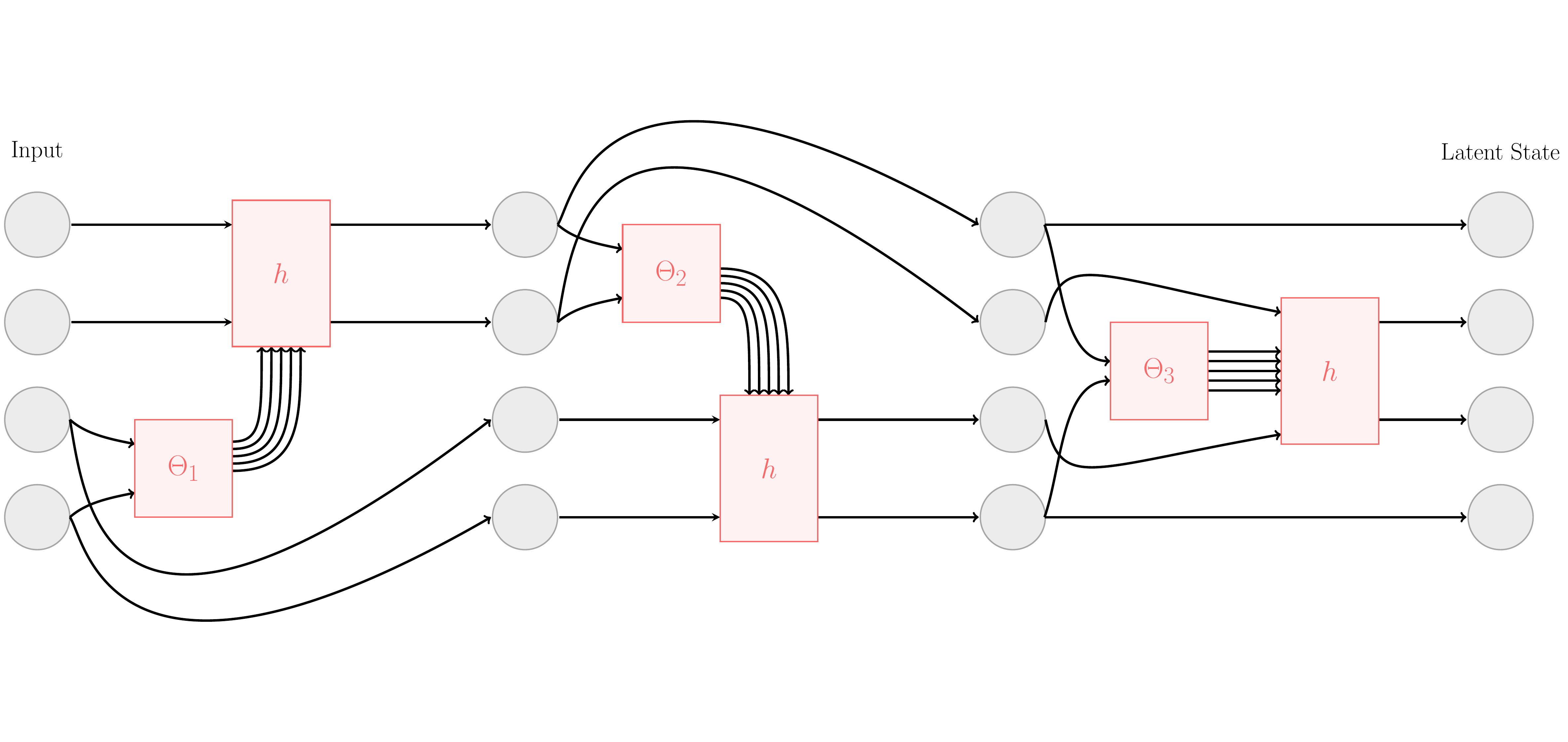}}
\caption{A visualization of the normalizing flow architecture used.} \label{fig2}
\end{figure*}

\section{Modeling Winds via Normalizing Flows}\label{appendix:NF}
The data at hand are often quite complex. As can be seen from Figure \ref{fig:reduced_data}, the response itself has quite a complex form, taking the form of a doughnut shaped probability distribution. The inherent complexity of the data is only exacerbated by the packing of high dimensional information into a low dimensional form. Because of this, quite expressive models must be used to capture the detail in these distributions. In the context of modeling via normalizing flows, such data demand the use of flexible, non-naive architectures modeling $\phi$, preferably in addition to expressive latent spaces.

Design of non-naive and expressive layers in the construction of normalizing flows is tricky given that invertibility must be guaranteed. The architectural component most heavily relied on in our modeling is what has been called the ``coupling flow'' in the literature. At the input to each coupling flow layer, an input $v \in \mathbb{R}^{p+m}$ is divided into two components $v^{A} \in \mathbb{R}^{r}$ and $v^{B} \in \mathbb{R}^{p+m-r}$.  The coupling flow transforms the input by pushing $v^A$ forwards by some invertible function $h$, and transforms $v^B$ by the identity transform. 

We define our coupling function $h: \mathbb{R}^r \to \mathbb{R}^r$ element-wise; let $\tilde{h}: \mathbb{R} \to \mathbb{R}$, $\tilde{h}_{\theta}(x) = ax + b + \frac{c}{1 + (dx + h)}$, $\theta = [a, b, c, d, g]$, and finally $h(v) = \big(\tilde{h}_{\theta_1}(v_1), \hdots, \tilde{h}_{\theta_r}(v_r)\big)$. We model $\theta_i$ as the transformed output of a shallow neural network $\Theta_i: \mathbb{R}^{p+m-r} \to \mathbb{R}^5$, $\Theta_i(v^B) = \theta_i'$, where specifically, $\theta_i$ is formed from $\theta_i' = [a', b', c', d', g']$ in the method of \citep{ziegler2019} to guarantee the invertibility of $h$:
\begin{equation*}
a = a', \ b = e^{b'}, \ c =  \frac{8 \sqrt{3}}{9d} \tanh(c'), \ d = e^{d'}, \  g = g'.
\end{equation*}
This particular coupling function $h$ is well-known in the literature, and has been noted in \citep{kobyzev2019} for it's particular expressive power.

So far, no fancy architecture has been used to model $\Theta_i$. As of now, we model  
\begin{equation*}
  \Theta_i \in \bigg\{ f: \mathbb{R}^{r} \to \mathbb{R}^5  \mid f = R \circ A^{(l)} \circ \hdots \circ R \circ A^{(1)} + L  \bigg\}, 
\end{equation*}
i.e. we learn a fully connected network with linear residual connection, and ReLu activations. The residual connections are suggestion of \citep{kingma2018}. We let $l$ range between 3 and 10 in our experiments, and $l \approx 5$ is the general default setting. The choice of $l$ does not appear to be terribly important as long as $l\geq 3$, which guarantees a flexible mapping. 

\begin{table*}[t]
\centering
\begin{tabular}{@{}rrrrcrrrcrrr@{}}
\toprule
\phantom{abc} & \textbf{Hyperparameter} & \textbf{Min. Value} & \textbf{Max. Value} &  \textbf{In Experiments}\\ \midrule

\textbf{Latent Distribution} \\
\phantom{abc} & \textbf{Choice} & {Gaussian Mixture} & {Gaussian Mixture} & {Gaussian Mixture} \\
\phantom{abc} & \textbf{Mixture Components} & {3} & {20} & {5} \\
\phantom{abc} & \textbf{Train Latent Dist. Params.} & {True} & {True} & {True} \\
\textbf{$\Theta$-Network} \\
\phantom{abc} & \textbf{$\Theta$-Net Depth} & {3} & {9} & {7} \\
\phantom{abc} & \textbf{$\Theta$-Net ResNet} & {False} & {True} & {True} \\
\textbf{Core Training Params.} \\
\phantom{abc} & \textbf{Training Iter.} & {200} & {4000} & {1200} \\
\phantom{abc} & \textbf{Batch Size} & {75} & {500} & {150} \\
\phantom{abc} & \textbf{Learning Rate} & {.01} & {.01} & {.01} \\
\phantom{abc} & \textbf{Optimizer} & {Adam} & {Adam} & {Adam} \\
\phantom{abc} & \textbf{$\beta_1$} & {.99} & {.99} & {.99} \\
\phantom{abc} & \textbf{$\beta_2$} & {.99} & {.99} & {.99} \\
\phantom{abc} & \textbf{Calibration Check Interval} & {50} & {200} & {100} \\
\bottomrule
\end{tabular}
\label{tab:tracker}
\caption{An overview of specific hyperparameter choices made.}
\end{table*}

The choice of which coordinates go into which component is a hyperparameter. As of now, it seems usually sufficient to have $v^A$ be the coordinates corresponding to response and $v^B$ be the coordinates corresponding to the predictors in one ``coupling layer'' and vice versa in a second coupling layer. These two make up the entire $\phi$ architecture. It is possible to stack further coupling flows on top of each other in the case that more model flexibility is required. 

The choice of latent distribution, i.e. the choice of $p_W$, is an even more
important hyperparameter. In practice, we have found it a useful addition to the
complexity of the model to let $p_W$ be the density of a normal mixture
\citep{izmailov2020semi}. As with all latent distributions, we treat the
parameters $\omega$ as trainable, so the only choice to be made by the user is
the number of mixture components. We have had success using anywhere between $1$
and $20$ mixture components, with more mixture components leading to more
complex models. Stacking many coupling flows on top of each other may lead to
numerical instabilities in the inversion of $\phi$, so this can be quite useful.
If more naive architectures are used to model $\phi$, using $20$ or more mixture
components can be a saving grace.

\end{document}